\documentclass[letterpaper]{article} 
\usepackage{aaai23}  
\usepackage{times}  
\usepackage{helvet}  
\usepackage{courier}  
\usepackage[hyphens]{url}  
\usepackage{graphicx} 
\usepackage{natbib}  
\usepackage{caption} 
\usepackage{algorithm}
\usepackage{algorithmic}

\usepackage{booktabs}
\usepackage{amsmath}
\usepackage{bm}
\usepackage{multirow}
\usepackage{lineno}

\usepackage{newfloat}
\usepackage{listings}
\DeclareCaptionStyle{ruled}{labelfont=normalfont,labelsep=colon,strut=off} 
\floatstyle{ruled}
\newfloat{listing}{tb}{lst}{}
\floatname{listing}{Listing}
\title{A Generative Approach for Script Event Prediction via Contrastive Fine-tuning}
\author{
    Fangqi Zhu\textsuperscript{\rm 1}\enskip
    Jun Gao\textsuperscript{\rm 1}\enskip
    Changlong Yu\enskip
    Wei Wang\enskip
    Chen Xu\textsuperscript{\rm 3}\\
    Xin Mu\textsuperscript{\rm 2}\enskip
    Min Yang\textsuperscript{\rm 4}\enskip
    Ruifeng Xu\textsuperscript{\rm 1,\rm 2,\rm 5}\thanks{Corresponding author}
}
\affiliations{
    \textsuperscript{\rm 1} \normalsize Harbin Institute of Technology (Shenzhen) \\
    \textsuperscript{\rm 2} \normalsize Peng Cheng Laboratory \quad
    \textsuperscript{\rm 3} \normalsize Beijing University of Technology \\
    \textsuperscript{\rm 4} \normalsize Shenzhen Institute of Advanced Technology,  Chinese Academy of Sciences\\
    \textsuperscript{\rm 5}Guangdong Provincial Key Laboratory of Novel Security Intelligence Technologies \\
    \texttt{\{zhufangqi.hitsz, imgaojun\}@gmail.com} \enskip \texttt{xuruifeng@hit.edu.cn}\\
}

\usepackage{bibentry}

\begin{document}
\maketitle 
\begin{abstract}
Script event prediction aims to predict the subsequent event given the context. This requires the capability to infer the correlations between events. Recent works have attempted to improve event correlation reasoning by using pretrained language models and incorporating external knowledge~(e.g., discourse relations). Though promising results have been achieved, some challenges still remain. First, the pretrained language models adopted by current works ignore event-level knowledge, resulting in an inability to capture the correlations between events well. Second, modeling correlations between events with discourse relations is limited because it can only capture explicit correlations between events with discourse markers, and cannot capture many implicit correlations. To this end, we propose a novel generative approach for this task, in which a pretrained language model is fine-tuned with an event-centric pretraining objective and predicts the next event within a generative paradigm. Specifically, we first introduce a novel event-level blank infilling strategy as the learning objective to inject event-level knowledge into the pretrained language model, and then design a likelihood-based contrastive loss for fine-tuning the generative model. Instead of using an additional prediction layer, we perform prediction by using sequence likelihoods generated by the generative model. Our approach models correlations between events in a soft way without any external knowledge. The likelihood-based prediction eliminates the need to use additional networks to make predictions and is somewhat interpretable since it scores each word in the event. Experimental results on the multi-choice narrative cloze~(MCNC) task demonstrate that our approach achieves better results than other state-of-the-art baselines. Our code will be available at \url{https://github.com/zhufq00/mcnc}.
\end{abstract}

\section{Introduction}
\begin{figure}[!t] 
    \centering
    \includegraphics[width=1.0\linewidth]{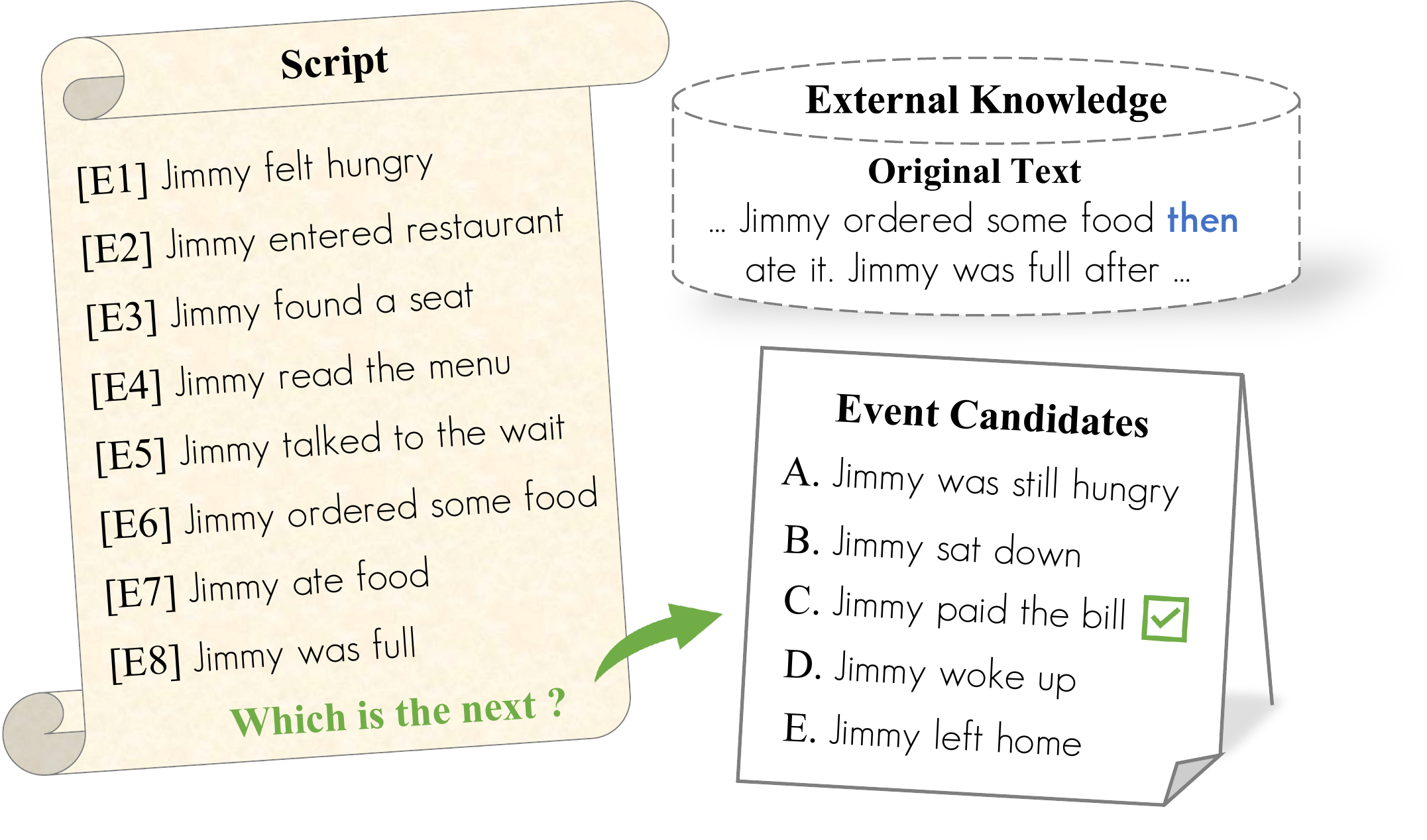}
    \caption{Script event prediction with external knowledge. Giving a script, the task aims to select the subsequent event from event candidates. Events E6$\sim$E8 are extracted from the original text where ``then'' is a discourse marker, which is used by \citet{lee-etal-2020-weakly} to extract the discourse relation. Note that our method is \textbf{free} with this kind of external knowledge (dashed box).}
    \label{fig:figure1}
\end{figure}

A script~\cite{schank2013scripts} refers to a kind of structured knowledge, which involves sequences of events.
Script event prediction aims at predicting the subsequent event given an event chain in a script.
Figure~\ref{fig:figure1} presents a restaurant script involving sequences of events that occurs when a person enters a restaurant. The goal of the script event prediction task~\cite{MarkGW-AAAI16} is to select the plausible subsequent event from the five event candidates. Solving the script event prediction problem can help us to acquire event knowledge from event chains and is beneficial for many downstream tasks such as story generation~\cite{chaturvedi-etal-2017-story} and dialogue generation~\cite{danescu-niculescu-mizil-lee-2011-chameleons}.

Understanding events and inferring correlations between events are crucial to acquiring event knowledge. Taking Figure~\ref{fig:figure1} for example, we first need to understand the events~(what is happening) and have the commonsense knowledge - ``You need to pay after the meal'', and then we can infer the plausible subsequent event - ``Jimmy paid the bill''~(what happens next). Early works~\citep{MarkGW-AAAI16,wang-etal-2017-integrating} used embedding method~(e.g., Word2Vec) to learn event representation and then choose the subsequent event by calculating similarities between event candidates and script events. 
After that, some works~\cite{lv-etal-2020-integrating} attempted to leverage masked language model~(MLM)~\citep{liu2019roberta} to obtain better event representations and they all achieved a substantial improvement. However, some studies~\citep{lee-etal-2020-weakly} have shown that these approaches based on pre-trained models still lack an understanding of the relationships between events.
Some recent works 
\citep{lee-etal-2020-weakly,bai-etal-2021-integrating} incorporate external knowledge to enhance the model to understand correlations between events, including discourse relations~\citep{bai-etal-2021-integrating} and original texts~\citep{lee-etal-2020-weakly}. \citet{lee-etal-2020-weakly}~(NG) extract discourse relations from original texts by a template matching-based method. \citet{bai-etal-2021-integrating} use original texts to enhance event representations~(See Figure~\ref{fig:figure1} for example).

Despite promising results achieved by these methods, some challenges still remain. 
First, the pretrained language models adopted by current works ignore event-level knowledge. Specifically, MLMs are pre-trained using token-level learning objectives and thus unable to capture the correlations between events. 
Second, modeling correlations between events with discourse relations is limited because it can only capture explicit correlations between events with discourse markers, and cannot capture many implicit correlations. For example, in Figure~\ref{fig:figure1}, a model can easily recognize a \textit{precedence} relation between E6 event and E7 event based on the discourse marker ``then'', while the model struggles to recognize the implicit event relation between E7 event and E8 event when there are no discourse markers.

To better model correlations between events and transfer the knowledge of pre-trained models~(including general knowledge and event knowledge) to the script event prediction task, we propose a novel generative approach for this task, in which a pretrained language model is fine-tuned with an event-centric pretraining objective and predicts next event within a generative paradigm. Specifically, our approach consists of two stages: event-centric pretraining stage and contrastive fine-tuning stage. (1) In the event-centric pretraining stage, We use the BART~\cite{lewis-etal-2020-bart} as the backbone for this task and introduce a novel event-level blank infilling strategy as the learning objective to inject event-level knowledge into the pretrained language model. We randomly mask several events in the script and then require the model to generate these masked events autoregressively. 
(2) In the contrastive fine-tuning stage, we design a likelihood-based contrastive loss to force the model to learn to distinguish between correct and wrong event candidates. To be specific, the model is required to reduce the generation probabilities of wrong event candidates and increase the generation probability of correct event candidates. 
Instead of using an additional prediction layer, we perform prediction by using sequence likelihoods generated by the generative model.
Our approach models correlations between events in a soft way without any external knowledge and maintain the consistency of the contrastive fine-tuning, event-centric pretraining, and pretraining objectives. 
Besides, the likelihood-based prediction eliminates the need to use additional networks to make predictions and is somewhat interpretable since it scores each word in the event. The main contributions of this work are summarized as follows:
\begin{itemize}
    \item We propose a novel generative approach for this task, in which a pretrained language model is fine-tuned with an event-centric pretraining objective and predicts the next event within a generative paradigm. 
    \item We introduce a novel event-level blank infilling strategy as the learning objective to inject event-level knowledge into the pretrained language model and design a likelihood-based contrastive loss to force the model to learn to distinguish between correct and wrong event candidates.
    \item Experimental results on the multi-choice narrative cloze~(MCNC) task demonstrate that our approach achieves better results than other state-of-the-art baselines.
\end{itemize}

\section{The Proposed Approach}
As shown in Figure~\ref{fig:figure1}, script event prediction can be defined as predicting the most probable subsequent event given a script~(an existing event chain). Formally, given a script $X=\left\{x_1,\dots,x_{n}\right\}$ and event candidates $Y=\left\{y_1,\dots,y_m\right\}$ where $x_i$ and $y_j$ represent event, this task aims to choose the correct subsequent event $y_t$ from $Y$. Following \citet{MarkGW-AAAI16}, each event $e=(e_v, e_s , e_o , e_i)$ consists of a predicate $e_v$ and three arguments: subject $e_s$, object $e_o$ and indirect object $e_i$. If an event does not have an argument, the argument will be denoted as \texttt{null}. Our proposed approach involves modeling a conditional distribution $P(Y|X)$. Our model should learn to calculate the conditional probability distribution $P(y_i|X)$ for each event candidate $y_i(i\in{1,\dots,m})$, given the script $X$. 

During training, our method consists of two stages: event-centric pretraining and task-specific contrastive fine-tuning. In the event-centric pretraining stage, we randomly mask some events in the script and then enable the model to generate the masked events sequentially. By doing so, our model models correlations between events and mines the event knowledge in the script. In the task-specific contrastive fine-tuning stage, we introduce wrong event candidates to make the model learn to distinguish between right and wrong event candidates. We train our model to maximize the generation probability of the right event candidate and reduce the generation probabilities of the wrong event candidates. During testing, our model calculates the generation probability of each event candidate and selects the one with the highest generation probability as the predicted subsequent event.

In recent years, pre-trained models like BART ~\citep{lewis-etal-2020-bart} have significantly improved over various downstream tasks such as question answering, summarization, and machine translation. Therefore, we adopt BART as the underlying architecture for our model to model the conditional probability distribution $P(Y|X)$.

\begin{figure*}[htbp] 
    \centering
    \includegraphics[width=1.0\linewidth]{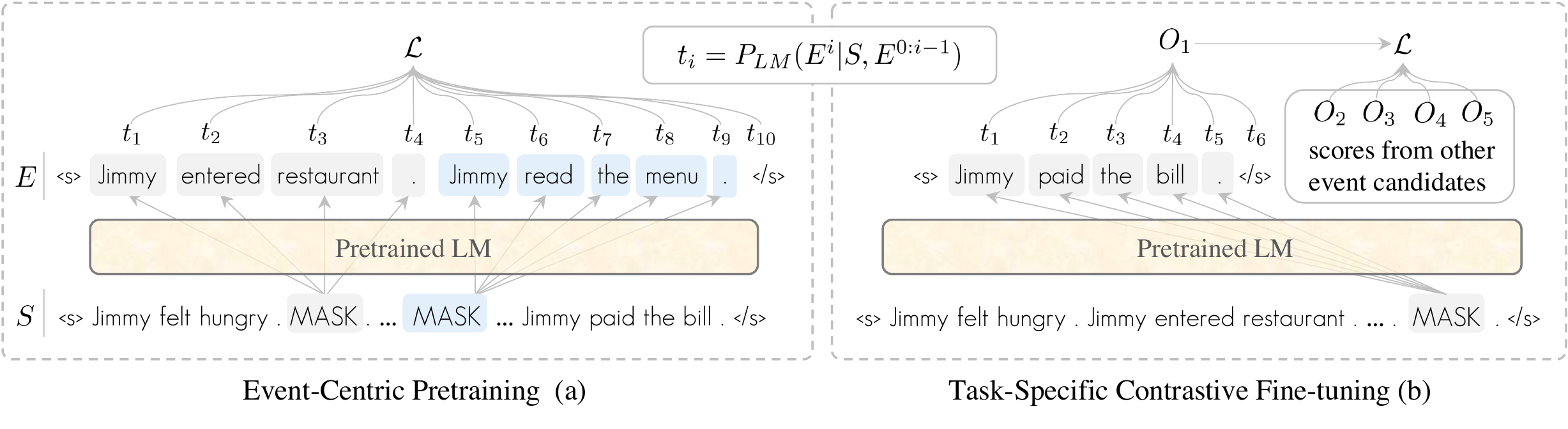}
    \caption{Our two-stage generative model training method. $S$ and $E$ correspond to the input and output of the generative model. $t_i$ corresponds to the probability of generating the token corresponding to $t_i$ given $S$ and $E^{0:i-1}$. In the task-specific contrastive fine-tuning stage, the model forwards five times to get the overall scores of the five event candidates, respectively.}
    \label{fig:figure2}
 \end{figure*}

\subsection{Event-Centric Pretraining}
In the event-centric pretraining stage, we aim to train our model to model correlations between events and mine the event knowledge in the script by an event-level blank infilling strategy. For a more complete usage of the script, we concatenate the script and the right event candidate $y_t$ to become the event sequence $S_\texttt{ori}=\left\{x_1,\dots,x_{n},y_t\right\}$. To enrich the variety of mask strategies and maintain the challenge of the event-centric pretraining stage, we randomly mask $K$ events in $S_\texttt{ori}$, where $K$ is randomly selected from 1 to 3. We denote the masked $S_\texttt{ori}$ as $S$ and the masked events as event sequence $E$ where the masked events are in the order they were in $S_\texttt{ori}$. Since the BART model requires natural language as input and output, as shown in Figure~\ref{fig:figure2}, we represent each event ${e}=(e_v,e_s,e_o,e_i)$ as ``$e_s\ e_v\ e_o\ e_i$' and the sequence of events $S$ as ``$\texttt{\textless s\textgreater} x_1 \texttt{\textless SEP\textgreater} x_2 \texttt{\textless SEP\textgreater} \texttt{\textless MASK\textgreater} \texttt{\textless SEP\textgreater} \dots \\\ \texttt{\textless SEP\textgreater} \texttt{\textless MASK\textgreater} \texttt{\textless SEP\textgreater} \dots \texttt{\textless SEP\textgreater} y_t \texttt{\textless \\s\textgreater}$'' where each masking event is replaced with a token \texttt{\textless MASK\textgreater} and each event is separated by \texttt{\textless SEP\textgreater}. Event sequence $E$ is represented similarly. Note that the first event $x_1$ and the last event $y_t$ may also be masked. Since the \texttt{\textless SEP\textgreater} token does not exist in the BART model, we use ``.'' as the separator. 
   
With the help of the autoregressive generative model BART, the conditional probability distribution $P(E|S)$ is formulated as follows:
\begin{equation}
 P(E|S) = \frac{1}{N_E}\sum_{n=2}^{N_{E}}{\log P_{LM}\left(E^{n}|S,E^{1:n-1}\right)}
\end{equation}
where $N_{E}$ is the token number of event sequence $E$ in natural language format, $E^{n}$ is the nth token, and $E^{1:n-1}$ is the first to n-1th token of $E$.

In the event-centric pretraining stage, we aim to train our model to maximize $P(E|S)$. We test the model on the development set and use the model parameters of the best epoch as initial parameters of the model in the next stage.

\subsection{Task-Specific Contrastive Fine-tuning}
In the task-specific contrastive fine-tuning stage, negative event candidates are introduced to support the generative model with discriminative ability. This step further enables the model to select the correct event from the event candidates, which is crucial to the final performance of the model. We first calculate the conditional generation probability of each event candidate as their scores. We then tried three different loss functions to optimize the scores so that the score of the right event candidate is higher than that of the negative event candidate in different ways.

More specifically, we first add a token \texttt{[mask]} at the end of the script $X$, denoting the modified sequence as $X_m$. Then $X_m$ and each event candidate $y_i \in Y$ are converted to a natural language format, as mentioned in Section 2.1. Since each event candidate has a different number of tokens, to avoid the model tending to choose shorter event candidates, we utilize the mean of log-probabilities of verbalizer tokens for each event as the score $o_i$ for the event $y_i$.
\begin{equation} \label{score}
 o_i = \frac{1}{N_{y_{i}}} \sum_{n=2}^{N_{y_{i}}}{\log P_{LM}\left(y_i^{n}|X_m,y_i^{1:n-1}\right)}
\end{equation}
where ${N_{y_{i}}}$ is the token number of the event $y_i$. Afterwards, we use \textrm{softmax} function to calculate the final score $s_i$ for each event candidate $y_i$:
\begin{equation}
 s_i = \frac{\exp{(o_i)}}{\sum_{k=1}^{M}\exp{(o_k)}}
\end{equation}
Finally, we define the loss function as follows:
\begin{equation}
    \mathcal L_{cot}=-\log (s_t) + \frac{1}{M-1} \sum_{\substack{ i=1 \\ i\neq t}}^{M}{(\frac{s_i}{1-s_t})\log (\frac{s_i}{1-s_t})}
\end{equation}
where $t$ is the subscript of the right event candidate $y_t$, and $M$ is the number of event candidates. The first loss term is the conventional softmax cross entropy objective, which aims to maximize the generation probability of the right event candidate. The second loss term takes inspiration from \citet{chen2018complement}, which can be understood as directly maximizing the entropies of the generation probabilities of negative events conditioned on the right event candidate $y_t$ not happening. The difference is that \citet{chen2018complement} uses two different optimizers to optimize the two loss terms, and our method performs better with pilot experiments.

Further, we explore two other loss functions to explore how to optimize the generation probability of event candidates better, including traditional cross entropy loss and margin ranking loss widely used in contrastive learning.\\
$\bullet$ \textbf{Cross Entropy Loss} is calculated as follows:
\begin{equation}
\mathcal L_{cross}=-\log (s_t)
\end{equation}
$\bullet$ \textbf{Margin Ranking Loss} is calculated as follows:
\begin{equation}
\mathcal L_{margin}=\!\!\!\!\!\!\!\sum_{1\le i\le M,i\neq t}{\!\!\!\!\!\!\!\max{(m-(s_i-s_t),0)}}
\end{equation}
where $m$ is a hyperparameter margin which controls the interval between the score of the right event candidate and the score of the negative event candidate. The loss will be zero when all $s_i$ minus $s_t$ is greater than the margin $m$.

During testing, we select the most probable event candidate with the highest score as the predicted subsequent event $y_p$ , where $ p = \mathop{\arg\max}\limits_{1\le i\le M}s_i$.
\section{Experiments}
In this section, we introduce the datasets, experimental setting and compared baselines.
Experimental results show our method achieves state-of-the-art performance on the multi-choice narrative cloze~(MCNC) task. We then perform ablation study and model training comparison to understand the effect of the model's key components and their variants on performance. Finally, we conduct case study to show how our model predicts the subsequent event. 

\subsection{Datasets}

In the literature on script event prediction tasks, there are two versions of benchmark datasets that are used to evaluate the methods.
For the original MCNC task, ~\citet{MarkGW-AAAI16} extracted event chains from the widely-used New York Times portion of the Gigaword corpus~\cite{graff2003english}. 
We used the released codes \footnote{\url{https://mark.granroth-wilding.co.uk/papers/what_happens_next/}} to reproduce the extraction pipeline, including pos tagging, dependency parsing, and coreference resolution, etc. 
After generating event chains, we manually filter the extremely short event chains and truncate the long event chain so that the length of the remaining event chain~(defined as \text{script}) is nine. The last event of the script is used for the positive event candidate. Negative event candidates are randomly sampled, where the protagonist is kept the same as the protagonist of the current script, and other arguments~(\textit{object} or \textit{indirect object}) are replaced randomly by other arguments from the same document. Finally, the dataset contains more than 1.4 million event chains, and we split the training, validation, and test sets as \citet{bai-etal-2021-integrating}. We denote it as the ``original dataset,'' and the statistics are shown in Table~\ref{table1}.

Due to the complex extraction pipeline and restricted access to Gigaword corpus, most of the existing work chooses the ``public dataset'' 
released by \citet{li2018constructing}. 
Compared with the original dataset, the public dataset only covers the event chains that include extracted event relations between them.
This filtering results in ten times smaller than the original one. 
We also follow the common practice of dataset split for training, validation, and testing in Table~\ref{table1}.
For both ``original'' and ``public'' datasets, each instance has five event candidates for both datasets, of which only one choice is correct. To demonstrate the effectiveness of our proposed approach, we choose to evaluate our model on both of them.

\begin{table}[htbp]
 \centering
 \small
 \begin{tabular}{lcc}
  \toprule
  & Original Dataset & Public Dataset \\
  \midrule
  Train set  & 1,440,295 & 140,331 \\
  Dev set & 10,000 & 10,000 \\
  Test set    & 10,000 & 10,000 \\
  \bottomrule
 \end{tabular}
 \caption{The statistics of the reproduced original dataset~\cite{MarkGW-AAAI16} and the public dataset~\cite{li2018constructing}.}
 \label{table1}
\end{table}

\subsection{Experimental Settings} 
Our proposed two-stage method includes event-centric pretraining and task-specific contrastive fine-tuning. To compare with baselines of different sizes of datasets, we conduct experiments on BART$_\texttt{base}$ and BART$_\texttt{large}$. For these two stages, the model is optimized by Adam~\cite{kingma2014adam}. The learning rate and weight decay are 1e-5 and 1e-6. Our model uses an early stop strategy to select the best epoch, with patience set to 5. For BART$_\texttt{base}$, the batch size is 256 and 64 in the two stages, respectively. For BART$_\texttt{large}$, the batch size is 128 and 32 in the two stages, respectively. For the task-specific contrastive fine-tuning stage, the learning rate is chosen from \{\underline{1e-5}, 2e-5, 3e-5\}. All the experiments are conducted on Tesla A100 GPU. We choose the model with the best result on the development set and report its result on the test set. Accuracy is adopted as the evaluation metric.

\subsection{Baselines}
For a comprehensive with the previous methods, we divided them into the following categories: 

\noindent \textbf{Event Representations:}  1)~\textbf{Event-Comp}~\cite{MarkGW-AAAI16} uses the training objectives like Word2Vec~\cite{mikolov2013efficient} to learn event embeddings and calculates pairwise similarities between script events and event candidates. 2)~\textbf{Pair-LSTM}~\cite{wang-etal-2017-integrating} uses LSTM to model the narrative order of script events. 3)~\textbf{SAM-Net}~\cite{Lv_Qian_Huang_Han_Hu_2019} uses LSTM and self-attention mechanism to capture diverse event segments. 4)~\textbf{MCPredictor}~\cite{bai-etal-2021-integrating} obtains event representations from pretrained Word2Vec and enhances them with original sentence representations obtained by pretrained BERT. Moreover, multiple similar event chains are utilized to aggregate script-level information to help select the subsequent event. 
5)~\textbf{SCPredictor-s} is an ablation of MCPredictor, removing additional similar scripts and the original sentence information. 
6)~\textbf{BERT-based SCPredictor-s} replaces Word2Vec with BERT for a fair comparison under a similar amount of parameters. 
7)~\textbf{BART}~\cite{lewis-etal-2020-bart} fine-tunes the pre-trained model BART with a linear classifier.
 
\noindent \textbf{Methods Enhanced with Structured Information:}
1)~\textbf{NG}~\cite{lee-etal-2020-weakly} extracts event relations from the original text of the script, then a narrative event graph is constructed based on the event relations. Finally, NG transfers this task to a link prediction task~\cite{chang-etal-2014-typed}. 
2)~\textbf{RoBERTa + Rep. Fusion}~\cite{lv-etal-2020-integrating} integrates external knowledge from eventuality knowledge graph, ASER~\cite{zhang2020aser} and uses RoBERTa for making prediction. 
3)~\textbf{RoBERTa + Know. Model}~\cite{zhou-etal-2021-modeling} learns a knowledge model from ASER to predict event relations. 4)~\textbf{SGNN}~\cite{li2018constructing} constructs a narrative event evolution graph via verb con-occurrence frequency to obtain more effective event representations. 5)~\textbf{SGNN+Int\&Senti} incorporates external intention and sentiment knowledge from ATOMIC~\cite{Sap_Le} into event representations.
6)~\textbf{GraphBERT}~\cite{du-etal-2022-graph} builds an event graph similar to SGNN and enhances BERT with the event graph.
\noindent \textbf{Event-centric Post-Pretraining:}
1)~\textbf{EventBERT}~\cite{zhou2022eventbert} pretrains RoBERTa on \textsc{BookCorpus}~\cite{zhu2015aligning} with three self-supervised contrastive learning objectives: \textit{Correlation-based Event Ranking}, \textit{Contradiction Event Tagging} and \textit{Discourse Relation Ranking}.
2)~\textbf{ClarET}~\cite{zhou-etal-2021-modeling} pretrains BART on \textsc{BookCorpus} with another three  self-supervised objectives: \textit{Whole Event Recovering}, \textit{Contrastive Event-correlation Encoding} and \textit{Prompt-based Event Locating}.
\begin{table}[tbp]
 \centering
 \small
 \begin{tabular}{lcc}
  \hline
   Methods & Acc. (\%)  & ext.\\
  \hline 
  \multicolumn{3}{l}{\textit{w/ external knowledge}} \\
  \hline
  SGNN + Int\&Senti & 56.03 & Int \& Senti\\
  RoBERTa$_\texttt{base}$ + Rep. Fusion
  & 58.66 & ASER\\
  RoBERTa$_\texttt{base}$ + Know. Model
  & 59.99 & ASER \\
  \hline
  \multicolumn{3}{l}{\textit{w/o external knowledge}} \\
  \hline 
  Random & 20.00 & w/o ext.\\
  Event-Comp & 49.57 & w/o ext. \\
  PairLSTM & 50.83 & w/o ext.\\
  SGNN & 52.45 & w/o ext.\\
  GraphBERT & \underline{60.72} & w/o ext. \\
  \hline
  BART$_\texttt{base}$ & 60.00 & w/o ext. \\ 
  \textbf{Ours}~(BART$_\texttt{base}$)  & \textbf{62.94} & w/o ext.\\
  \hline
 \end{tabular}
 \caption{Base model accuracy on the test set of the public dataset. \textit{ext.} is short for \textit{external knowledge}.} 
 \label{table2}
\end{table}

\begin{table}[tbp]
 \centering
 \small
 \begin{tabular}{lcc}
  \hline
   Methods & Acc. (\%)  & ext.\\
  \hline
  \multicolumn{3}{l}{\textit{w/ external knowledge}} \\
  \hline
  EventBERT & 63.50 & \textsc{BookCorpus}\\
  RoBERTa$_\texttt{large}$ + Know. Model & 63.62 & ASER\\
  ClarET & 64.61 & \textsc{BookCorpus}\\
  \hline
  \multicolumn{3}{l}{\textit{w/o external knowledge}} \\
  \hline
  \textbf{Ours}~(BART$_\texttt{large}$)  & \underline{64.82} & w/o ext.\\
  \textbf{Ours}~(BART$_\texttt{large}$) + NYT & \textbf{65.88} & NYT \\
  \hline
 \end{tabular}
 \caption{Large model accuracy on the test set of the public dataset. NYT denotes the New York Time portion of the Gigaword corpus.}
 \label{table3}
\end{table}

\subsection{Results and Analyses}

We first present the main experimental results of the widely-used ``public'' dataset in Table~\ref{table2} and Table~\ref{table3} using the \texttt{base} and \texttt{large} model, respectively, in order to align with the existing baselines for comparable parameters. We can draw the following observations from the results on two model settings: 1). our approach achieves the new state-of-the-art performance with a comparable amount of parameters and obtains an absolute 2.22\% over the best baseline \textit{GraphBERT} without any external knowledge. To reduce the sparsity of the constructed event graph, GraphBERT uses the verb bigrams' frequency of two events to model the weights of event relations. By doing so, GraphBERT can only model correlations between verbs and ignore the event arguments. 
By contrast, our generative approach enables our model to capture correlations among all the events in one event chain. 
2). Although directly fine-tuning BART achieves much stronger performance than BERT or RoBERTa models, our approach further improves and increases the accuracy up to 2.94\% compared to vanilla BART$_\texttt{base}$.
3). Moreover, our approach even outperforms strong baselines that perform heavy event-centric post-pretraining such as ClarET~\cite{zhou-etal-2022-claret} and EventBERT~\cite{zhou2022eventbert}. Note that our approach(except Ours~(BART$_\texttt{large}$) + NYT) requires no external knowledge~(either from the corpus itself or eventuality knowledge graph like ASER~\cite{zhang2020aser}). 
4). To push the boundary of our proposed approach and test whether more resources can further enhance our approach, we replace the training set of the ``public dataset'' with the training set of the ``original datasets'' and then evaluate on the test set of the ``public dataset''. The result~({Ours~(BART$_\texttt{large}$) + NYT} in Table~\ref{table3}) demonstrates that more extensive training data can learn more precise event knowledge and final gains 1.27\% improvement over the previous state-of-the-art model ClarET. 
Finally, it is worth mentioning that our event-centric pretraining stage only needs 3.2 GPU hours on 0.76M tokens, while ClarET needs 90 GPU hours at 200M tokens~(external Bookcorpus). 
It significantly reduces the training time and requires much fewer computational resources, further demonstrating the efficiency of our approach. 

\begin{table}[tbp]
 \centering
 \small
 \begin{tabular}{llc}
  \hline
   Methods & Acc. (\%)  & ext. \\
  \hline 
  \multicolumn{3}{l}{\textit{w/ external knowledge}} \\
  \hline
  NG & 63.59 & discourse relation\\
  MCPredictor & \underline{67.14}* & original text\\
  \hline
  \multicolumn{3}{l}{\textit{w/o external knowledge}} \\
  \hline 
  SAM-Net & 55.60 & w/o ext. \\
  SCPredictor-s & 58.79* & w/o ext.\\
  BERT-based SCPredictor-s & 59.13 & w/o ext.\\
  \hline
  \textbf{Ours}~(BART$_\texttt{base}$)  & \textbf{67.21} & w/o ext.\\
  \hline
 \end{tabular}
 \caption{Model accuracy on the test set of the original dataset. ``*" indicate our reproduced results on our preprocessed original dataset.}
 \label{table4}
\end{table}

Then we evaluate our approach on the original dataset, show the experimental results in Table~\ref{table4} and observe the following key findings. 
1). Our method still achieves comparable performance to MCPredictor, although MCPredictor improves performance by 8.35\% with additional original sentence and multi-script knowledge aggregation. As we know, the original sentences where the events are extracted are critical during the training. However, the original sentences required by MCPredictor may not exist during testing. We tried to use the BM25 algorithm to retrieve the original text in the training set according to the test set's events to replace the original text in the test set. The accuracy dropped from 67.14\% to 50.27\%, which proves that the original text is indispensable for MCPredictor. At the same time, MCPredictor needs to re-generate additional scripts from original text based on event candidates, which is also inconvenient for applying the model to downstream tasks. 
2). To compare with MCPredictor with comparable parameters and training data, we tried changing the event encoder in SCPredictor-s from Word2Vec to BERT$_\texttt{base}$, and the experimental result shows similar performance. 
In summary, our approach can achieve comparable performance to MCPredictor without external knowledge, which benefits transferring the model to downstream tasks.

\begin{table}[tbp]
 \small
 \centering
 \begin{tabular}{lc}
  \toprule
   Methods & Acc. (\%) \\
  \midrule
  \textbf{Ours}~(BART$_\texttt{base}$)  & 62.94 \\
  \midrule
    w/o event-centric pretraining  & 61.08 \\
    w/o task-specific contrastive fine-tuning & 40.00 \\
    replace with a linear classifier & 61.77 \\
    replace with random span mask & 61.44 \\
    replace with sum of log-probabilities &  60.84 \\
    replace with Cross Entroy Loss &  62.71 \\
    replace with Margin Ranking Loss & 61.18 \\

  \bottomrule
 \end{tabular}
 \caption{Ablation study of our BART$_\texttt{base}$ model accuracy on the test set of the public dataset.}
 \label{table5}
\end{table}

\subsection{Ablation Study} 
Table~\ref{table5} shows our ablation analysis for our approach. We first investigated the impact of the event-centric pretraining stage~(row1) and found that the model's performance dropped by 1.86\% without the stage. This is because the model captures correlations between events in the stage, enabling the model to reason about unseen events based on a partial event sequence. 
Second, we remove our task-specific contrastive fine-tuning stage~(row2) or switch to the traditional classification method as in BART~(row3). We can observe that the task-specific contrastive fine-tuning stage is crucial because it enables the model to distinguish between correct and incorrect event candidates. 
Moreover, it is better to fine-tune the model in our generative method instead of the discriminative classification method, because our method minimizes the inconsistency between task-specific contrastive fine-tuning, event-centric pretraining and BART pretraining stages. We then conducted experiments with the random span masking strategy to verify the effectiveness of the event-level blank infilling strategy. The only difference is that the masked span is no longer a complete event~(the length and number of the masked spans keep unchanged). We find that the model's performance is deficient in the event-centric pretraining stage~(35.30\%), and the event knowledge in the script is not well learned, which leads to its final performance drops~(row4). Another important observation is that optimizing the mean of log-probabilities ~(Eq. \ref{score} in our method) is significantly better than optimizing the sum of log-probabilities ~(row5). The latter will tend to select shorter event candidates since they have fewer log-probability items. Finally, we explore the performance of the two variant losses. We find that our introduced \textit{ComplementEntropyLoss}'s performance is slightly better than the traditional \textit{Cross Entropy Loss}~(row6) and much better than the \textit{Margin Ranking Loss}~(row7), which is widely used in contrastive learning.

\begin{figure}[tbp] 
    \centering
    \includegraphics[width=0.8\linewidth]{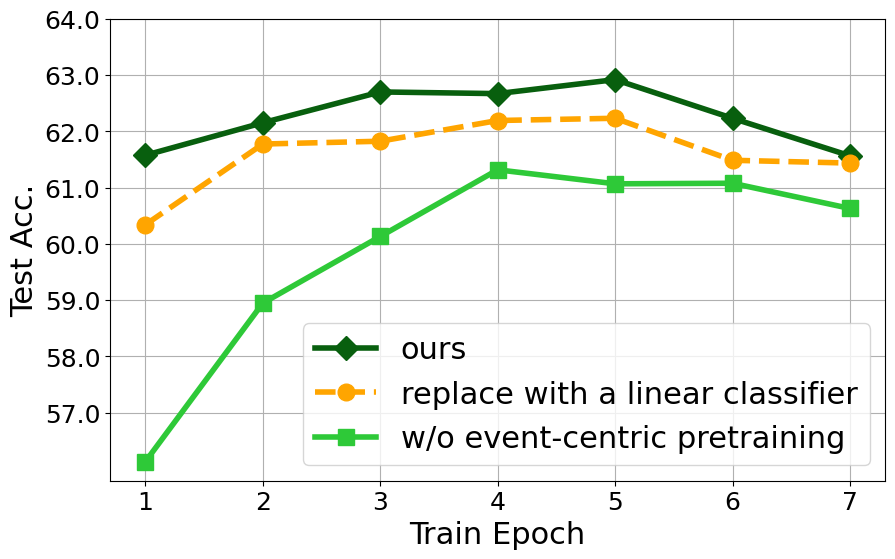}
    \caption{The training process comparison between ours and two ablations~(\textit{w/o event-centric pretraining} and \textit{replaced with a linear classifier}). We test our model on the test set at each epoch.}
    \label{fig:figure3}
\end{figure}

\subsection{Model Training Comparison}
In this section, we compare the training process on the public dataset of Ours~(BART$_\texttt{base}$) with two ablation methods~(\textit{w/o event-centric pretraining} and \textit{replaced with a linear classifier}). From Figure~\ref{fig:figure3}, we can see that our method achieves higher results than the two ablation methods at every epoch. Compared with \textit{w/o event-centric pretraining}, our method leads by a large margin in the first epoch because our method models correlations between events and learns event knowledge. Note that the model has not seen the data in the test set during the event-centric pretraining stage, so the event knowledge is generalizable. Compared with \textit{replaced with a linear classifier}, our model leads by a more considerable margin at the initial epoch than in the later epochs, due to a more negligible difference between our task-specific contrastive fine-tuning and event-centric pretraining and pretraining objectives.

\begin{figure}[htbp] 
    \centering
    \includegraphics[width=0.9\linewidth]{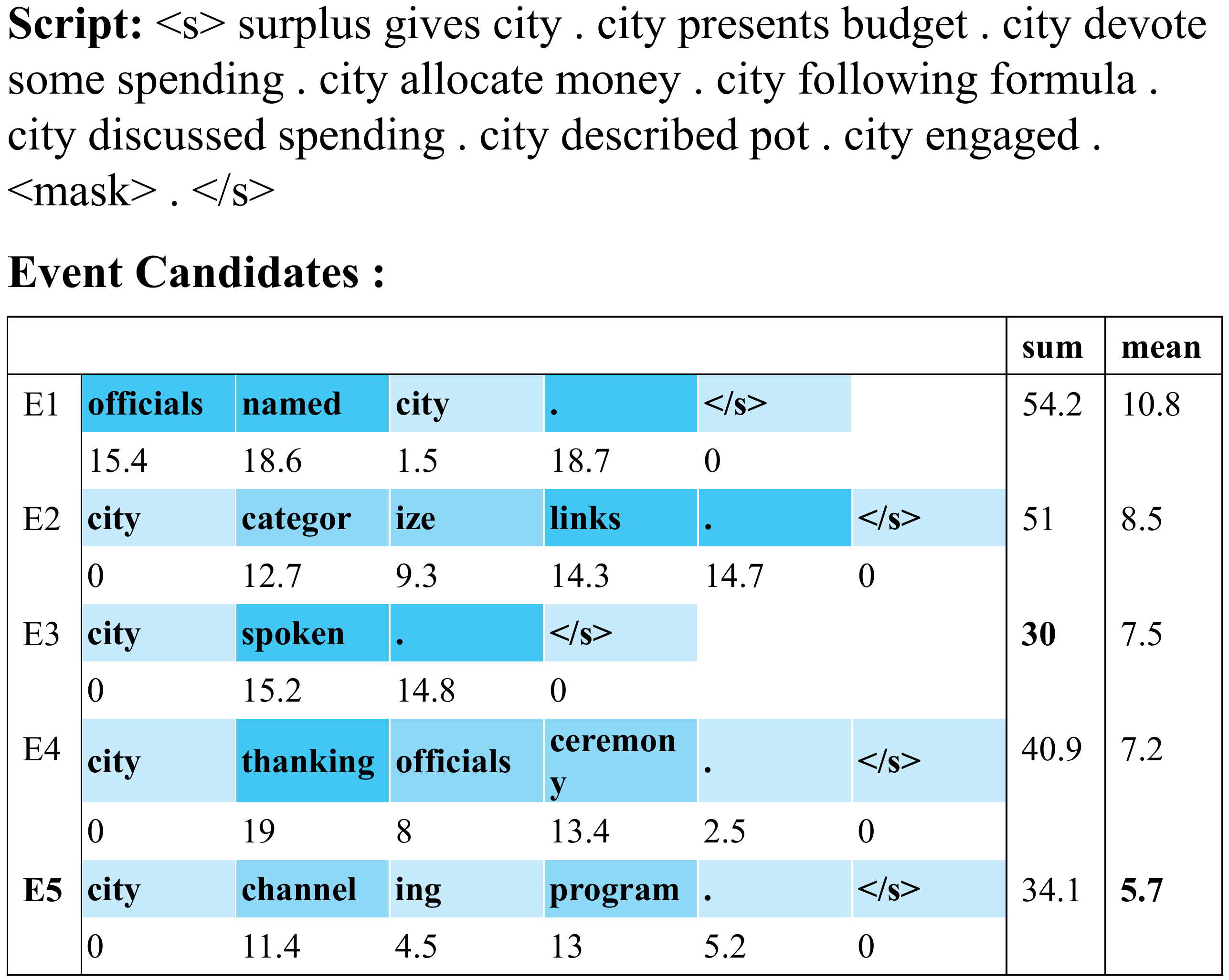}
    \caption{Case Study. The script corresponds to the input of our model in the contrastive stage. Below the token of the event candidate is its negative log generation probability. The higher the value, the harder it is to generate the token. \textbf{E5} is the correct event candidate.}
    \label{fig:figure4}
\end{figure}

\subsection{Case Study}
We conduct a case study in the test dataset to show some properties of our model and illustrate how our model predicts the correct answer. First, we notice that the model can easily predict that the event candidate's first token is ``city'' because ``city'' is the script's protagonist. And the event candidate's protagonist should be consistent with the script's protagonist. In addition, the model can also perceive that ``.'' must be followed by \textit{\textless /s \textgreater}. Both of these show that the model can grasp some simple rules. Second, our model generates the score of each token in the event candidate in an autoregressive way. Our model is also interpretable to a certain extent. For example, the scores of ``.'' in E1, E2, and E3 are all high, mainly because the model thinks the event is not over after generating the front part of the event. However, for E4 and E5, the model focuses more on the plausibility of verbs and objects. Finally, this case shows how our model avoids favoring shorter event candidates. In this case, E3 has the smallest sum of negative log generation probabilities, while E5 has the smallest mean of negative log generation probabilities. Compared with the ablation method~(\textit{replace with sum of log-probabilities}), our model can combine the length of events to select a more reasonable event candidate.

\section{Related Work}
Script event prediction was started by \citet{chambers-jurafsky-2008-unsupervised}, who proposed the \textit{narrative cloze} task: remove an event from a narrative event chain and adopt the model to predict the missing event. \citet{chambers-jurafsky-2008-unsupervised} define an event as a verb and its dependency, and predict the missing event by calculating the frequency of co-occurrence between events. Afterward, \citeauthor{MarkGW-AAAI16} \shortcite{MarkGW-AAAI16} expand the definition of an event as a verb and its three arguments (subject, object, indirect object) and propose the widely used \textit{multiple choice narrative cloze}~(MCNC) task: select the subsequent event from event candidates according to the historical narrative event chain, which can also be denoted as script. Compared with the narrative cloze task, MCNC is more suitable for evaluating the effectiveness of script event prediction models. Subsequent works are almost carried out based on MCNC.

Early work~\cite{MarkGW-AAAI16} obtains the event representations through Word2Vec, then aggregates the pairwise similarities between an event candidate and script events to infer the probability that the event candidate is the subsequent event of the script. 
However, this method ignores the narrative order of script events, and subsequent works~\cite{wang-etal-2017-integrating,Lv_Qian_Huang_Han_Hu_2019} capture the narrative order through LSTM. Latter methods are more concerned with understanding event relations, which can be divided into two lines. 
The first line treats the verb co-occurrence frequency as an edge. Using this approach, \citeauthor{li2018constructing} \shortcite{li2018constructing} constructs a narrative event graph, and adopts a scaled graph neural network to obtain better event representations. Following them, \citet{ding-etal-2019-event-representation} enriches event representations with external intention and sentiment knowledge. 
\citet{gao2022improving} introduce a simultaneous weakly supervised contrastive learning and clustering method for event representation learning.
\citet{du-etal-2022-graph} introduces the pre-trained model BERT~\cite{devlin-etal-2019-bert} and replaces the middle layer of the model with a graph neural network to embed the event graph information. 
The second line introduces event relations to help the model select the subsequent event. \citet{lee-etal-2020-weakly} mine the discourse relations from the original text by template matching method, then construct a narrative event graph, and finally transform the script event prediction task into a link prediction task on the graph. \citet{bai-etal-2021-integrating} enriches the event representations with original text. \citet{lv-etal-2020-integrating,zhou-etal-2021-modeling} introduce eventuality knowledge graph ASER to enhance pre-trained model RoBERTa~\cite{liu2019roberta}.
However, the second line methods introducing the discourse relation have narrow discourse relation types and cannot capture various implicit correlations. Meanwhile, to avoid the sparsity of the event graph, the first line methods model the co-occurrence frequency of verbs, ignoring other event arguments. Our method can effectively model correlations between events through the event mask strategy.

Recent event pretraining studies~\cite{zhou2022eventbert,zhou-etal-2022-claret} conduct event-centric pretraining on \textsc{BookCorpus}. \citet{zhou-etal-2022-claret} and our method both adopt an event-level mask strategy. The difference is that our method masks more events, and we remove the other two pretraining objectives to focus on modeling event correlation. At the same time, our method uses the scripts in the training set for pretraining, and the learned event knowledge is closer to the MCNC task, making our model more powerful and efficient.
Besides,  the current methods introducing pretraining models use traditional classification fine-tuning methods when fine-tuning the models, resulting in inconsistency between pretraining and fine-tuning. Our method maximizes the consistency between pretraining and fine-tuning, enabling the model to better utilize the knowledge in the pre-trained model.

\section{Conclusion}
In this paper, we propose a novel generative approach, consisting of two stages, for the script event prediction task.
For the event-centric pretraining stage, we introduce a novel event-level blank infilling strategy as the learning objective to inject event-level knowledge into the pretrained language model. 
In the task-specific contrastive fine-tuning stage, we design a likelihood-based contrastive loss to force the model to learn to distinguish between correct and wrong event candidates. Experimental results on the multi-choice narrative cloze~(MCNC) task demonstrate that our approach achieves better results than other state-of-the-art baselines that require heavy pretraining or external knowledge. 
We believe our proposed approach can be a powerful ranking component for downstream tasks such as dialogue generation and story generation.

\section{Acknowledgments}
This work was partially supported by the National Natural Science Foundation of China (62006062, 62176076, 62106114, 61906185), the Shenzhen Foundational Research Funding (JCYJ20210324115614039, JCYJ20200109113441941), Key Technologies Research and Development Program of Shenzhen JSGG20210802154400001, the Major Key Project of PCL2021A06, Guangdong Provincial Key Laboratory of Novel Security Intelligence Technologies 2022B1212010005k.
\bibliography{aaai23}
\end{document}